\pdfoutput=1
\PassOptionsToPackage{prologue,dvipsnames}{xcolor}
\documentclass[11pt]{article}

\usepackage[preprint]{acl}
\usepackage{makecell}
\usepackage{times}
\usepackage{latexsym}
\usepackage{colortbl}
\usepackage[T1]{fontenc}
\usepackage{amssymb} 

\usepackage[utf8]{inputenc}

\usepackage{microtype}

\usepackage{inconsolata}

\usepackage{graphicx}
\usepackage{listings}
\usepackage[dvipsnames]{xcolor}
\usepackage{multirow}
\usepackage{booktabs}
\usepackage{amssymb}  
\usepackage{fontawesome5}
\usepackage[acronym,toc,xindy]{glossaries} 
\usepackage{tcolorbox}

\usepackage{float}
\usepackage[utf8]{inputenc} 
\usepackage{geometry} 
\usepackage{tikz} 
\usepackage{lipsum}

\newcommand{\mybox}[4]{
    \par 
    \noindent 
    \begin{tikzpicture}
        \node[anchor=text,text width=\columnwidth-1.2cm, draw, rounded corners, line width=1pt, fill=#3, inner sep=5mm] (big) {#4}; 
        \node[draw, rounded corners, line width=.5pt, fill=#2, anchor=west, xshift=5mm] (small) at (big.north west) {#1};
    \end{tikzpicture}
    \par 
    \vspace{5pt} 
}

\title{LogicQA: Logical Anomaly Detection with \\ Vision Language Model Generated Questions}

\author{Yejin Kwon$^{*}$, Daeun Moon$^{*}$, Youngje Oh, Hyunsoo Yoon$^{\dagger}$  \\
  Department of Industrial Engineering, Yonsei University \\
  Seoul, South Korea \\ 
  \texttt{\footnotesize\{beckykwon, dani0403, yj89.oh, hs.yoon\}@yonsei.ac.kr}}

\begin{document}
\maketitle
\begin{abstract}
Anomaly Detection (AD) focuses on detecting samples that differ from the standard pattern, making it a vital tool in process control. Logical anomalies may appear visually normal yet violate predefined constraints on object presence, arrangement, or quantity, depending on reasoning and explainability. We introduce LogicQA, a framework that enhances AD by providing industrial operators with explanations for logical anomalies. LogicQA compiles automatically generated questions into a checklist and collects responses to identify violations of logical constraints. LogicQA is training-free, annotation-free, and operates in a few-shot setting. We achieve state-of-the-art (SOTA) Logical AD performance on the public benchmark, MVTec LOCO AD, with an AUROC of 87.6\% and an $F_1$-max of 87.0\% along with the explanations of anomalies. Also, our approach has shown outstanding performance on semiconductor SEM corporate data, further validating its effectiveness in industrial applications.
\end{abstract}

\section{Introduction}
Anomaly detection (AD) is crucial for quality control and process optimization in industrial manufacturing. Anomalies are categorized into structural anomalies, referring to localized defects such as deformation or contamination \citep{bergmann2022beyond, 10758311}, and logical anomalies, which assess adherence to predefined constraints, including object presence, quantity, and arrangement \citep{batzner2024efficientad, kim2024few}. Unlike structural anomalies, logical anomalies demand clear explanations, as lack of reasoning may lead to misinterpretation. This necessitates an approach that not only detects but also explains logical anomalies \citep{zhang2024logicode}.

Data-driven AD plays a critical role in high-quality production and minimizing downtime in industrial control systems. However, simply detecting anomalies without explanation is insufficient \citep{wang2018sensor}. Modern industrial systems demand explainability to clarify the reasons behind anomalies \citep{li2023survey, gramelt2024interactive}. Understanding root causes enables security experts to take targeted actions, preventing severe malfunctions and unplanned stoppage \citep{10443463}.

\begin{figure}[t]
  \includegraphics[width=\columnwidth]{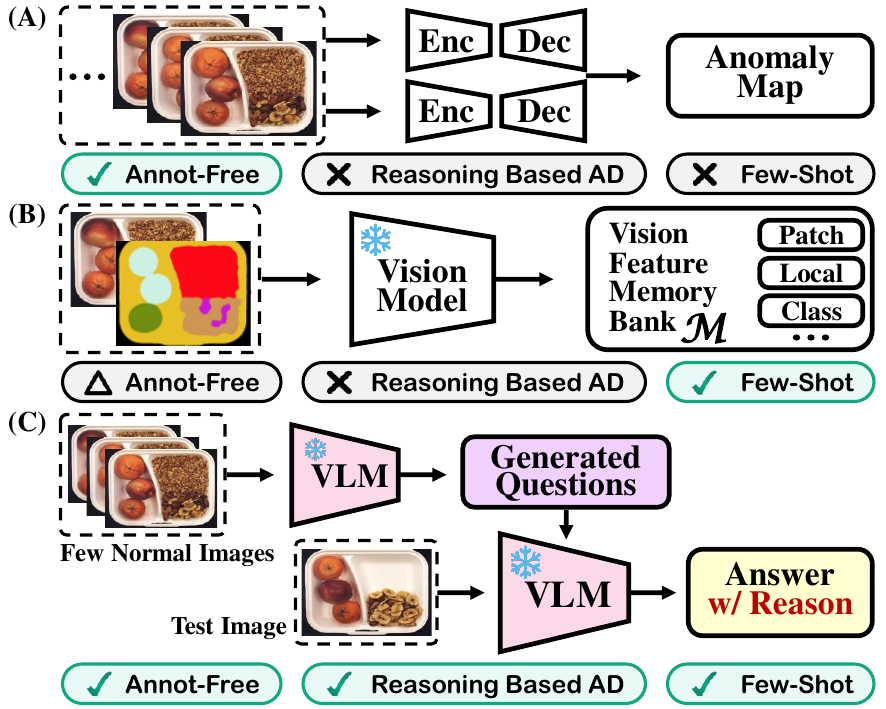}
  \caption{\textbf{Overview of Logical AD}: (A) Models trained from scratch (e.g., AutoEncoder) perform logical AD but require a large number of images. (B) Models leveraging memory-based AD methods (e.g., PatchCore) use pre-trained vision models to extract visual features from normal images, enabling few-shot AD. (C) Our method, LogicQA, utilizes a pre-trained VLM to generate anomaly-relevant questions and analyze test images, using the answers to identify and explain abnormalities.}
  \label{fig:overview_of_ad}
\end{figure}

Existing AD scores, estimating the probability of an image being anomalous, offer limited interpretability regarding the cause of anomalies \citep{sipple2022general}. As shown in Figure~\ref{fig:overview_of_ad}(A) and (B), most approaches rely on anomaly maps derived from pixel-wise anomaly scores \citep{tien2023revisiting, hsieh2024csad, Liu_2023_CVPR}. These heatmaps highlight abnormal regions but fail to explain why an anomaly has occurred. \textbf{LogicQA} (Logical Question Answering) (Figure~\ref{fig:overview_of_ad}(C)) addresses this limitation by leveraging a Vision-Language Model (VLM) to generate anomaly-relevant questions and provide natural language explanations, enhancing human interpretability. 

\textbf{LogicQA} introduces a few-shot logical AD framework leveraging a pre-trained VLM. Unlike conventional methods requiring class-specific models, LogicQA eliminates the need for training and manual annotations, allowing universal applicability across different classes. With just few normal images, LogicQA efficiently detects anomalies, making it scalable and practical for industrial fields. 

We validate \textbf{LogicQA} on the MVTec LOCO AD dataset \citep{bergmann2022beyond} and real-world semiconductor SEM dataset. This evaluation demonstrates its effectiveness in AD, particularly in semiconductor defect detection, and highlights its potential for broader industrial AD applications.

Our key contributions are as follows: (1) We achieve SOTA performance in few-shot logical AD by proposing LogicQA, using a VLM to generate anomaly-relevant questions and detect anomalies through question answering. (2) We enhance explainability in logical AD by generating natural language reasoning, helping engineers understand why logical anomalies occur. (3) We introduce a training-free and annotation-free approach, eliminating class-specific training and human-generated prompts, enabling efficient AD with few normal images for industrial uses. (4) We validate LogicQA on both public benchmark and real-world semiconductor SEM data, demonstrating its effectiveness across diverse AD settings.

\section{Related Work}
\paragraph{Logical AD Approaches}
Since the release of the MVTec LOCO AD dataset \citep{bergmann2022beyond}, various unsupervised AD approaches have been developed. Reconstruction-based methods \citep{bergmann2022beyond, An2015VariationalAB} rely on AutoEncoders trained with large amounts of normal images, limiting their applicability in few-shot scenarios. As PatchCore \citep{roth2022towards} was introduced, vision memory bank-based methods \citep{kim2024few, liu2023component} leverage pre-trained vision models and feature banks to improve efficiency. However, these methods require costly computational resources for fine-tuning. In contrast, LogicQA enables logical AD without fine-tuning, making it more scalable and adaptable to real-world applications.

\paragraph{VLMs for Logical AD}
Recent advancements in VLMs have enabled more interpretable AD by integrating vision and natural language reasoning \citep{achiam2023gpt, liu2024improved}. LogicAD \citep{jin2025logicad} employs a pre-trained VLM as a text feature extractor, generating explanations via logical reasoning. However, it relies on class-specific Guided Chain-of-Thought (CoT) prompts, requiring precise and laborious prompt engineering for each anomaly category. Similarly, LogiCode \citep{zhang2024logicode} applies Large Language Models (LLMs) to generate Python-based logical constraints, achieving strong detection performance but relying on detailed manual annotations, restricting practical industrial scalability. Our LogicQA overcomes these limitations by eliminating the need for pre-defined prompts and manual annotations, making it a more efficient and adaptable solution for industrial AD.

\section{LogicQA}
Logical AD differs from structural AD in that it assesses whether an image adheres to predefined logical constraints rather than identifying localized defects. Since logical anomalies often appear visually normal, detecting violations requires an interpretable framework to explain the underlying reasoning. 

\begin{figure*}[t]
    \centering
    \includegraphics[width=\textwidth]{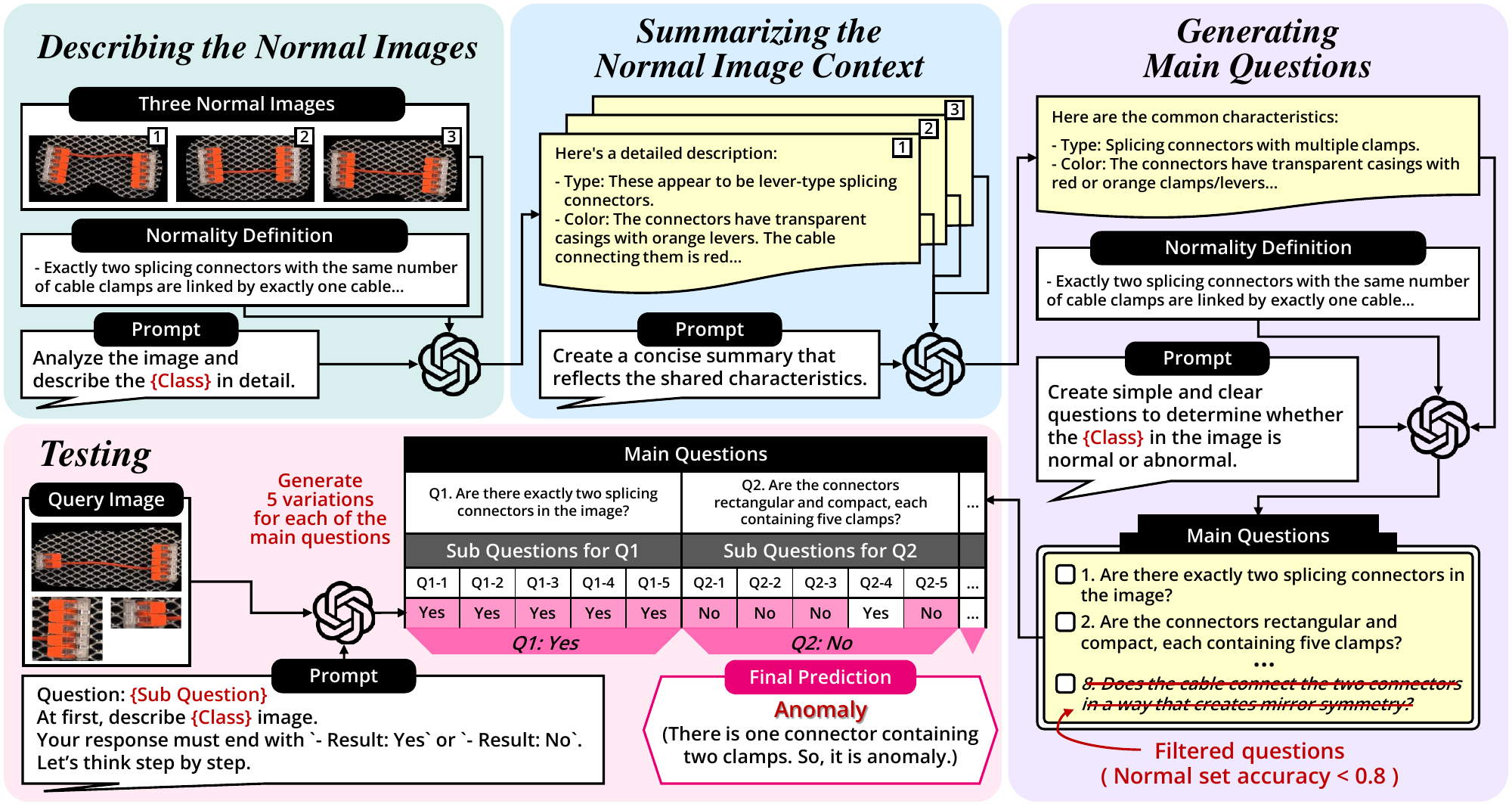}
    \caption{\textbf{Pipeline of LogicQA.} (1) \textbf{Describing the Normal Images} – The VLM generates textual descriptions of three normal images based on a predefined normality definition. (2) \textbf{Summarizing the Normal Image Context} – Shared features are extracted to define the core traits of normality. (3) \textbf{Generating Main Questions} – The VLM formulates key questions to assess whether an image is normal or anomalous. (4) \textbf{Testing} – The VLM generates sub-questions as variations of the main questions. Using a voting mechanism on the VLM’s responses, we determine whether the image satisfies the main questions. If it fails to satisfy even one, it is classified as anomalous.}
    \label{fig:main_figure} 
\end{figure*}

\subsection{Framework Overview}
\textbf{LogicQA} (Logical Question Answering) is a novel framework for logical AD that ensures interpretability by generating anomaly-relevant questions and reasoning. Unlike prior methods dependent on manual annotations or class-specific prompts, LogicQA leverages a pre-trained VLM, eliminating the need for annotations and human-generated prompts. This enables scalable deployment in industrial applications without task-specific fine-tuning.

Our proposed LogicQA consists of four stages: (1)\textbf{\textit{ Describing the normal images}}, (2)\textbf{\textit{ Summarizing the normal image context}}, (3)\textbf{\textit{ Generating the main questions}}, and (4)\textbf{\textit{ Testing}}, as shown in Figure~\ref{fig:main_figure}. All detailed prompts and examples are listed in the Appendix~\ref{sec:logicqa_prompt}.

\subsection{Describing the Normal Images} 
To ensure effective logical AD, LogicQA begins by analyzing the characteristics of normal images using a pretrained VLM. A single normal image, along with a predefined normality definition, is fed to the model, prompting it to generate a detailed textual description \citep{jin2025logicad}. The normality definition, adopted from \citet{bergmann2022beyond} (Appendix~\ref{subsec:normal_df}), establishes logical constraints that define expected object attributes and configurations in the dataset.

The descriptions capture location, quantity, and appearance of key elements, ensuring that the model focuses on relevant structural and contextual features rather than background noise. This process enhances AD robustness by aligning the model’s attention with critical aspects of normality. To further refine the understanding of normality, three distinct normal images are processed separately, with each description contributing to a consolidated representation of the dataset’s normality definition. This enables the model to generalize beyond individual examples, preserving essential normal properties.

\subsection{Summarizing the Normal Image Context}
The summarization step refines the extracted normality by feeding previously generated descriptions into the VLM and distilling shared attributes into a coherent representation of common features. This process ensures that AD remains robust against variations within normal images by focusing on the most consistent and core characteristics.

By using diverse normal images, the model learns robust normality patterns, ensuring AD remains effective across different instances. This prevents overfitting to specific examples and allows model to focus on meaningful logical constraints.

\subsection{Generating Main Questions} 
The question generation step refines generalized normality criteria into a checklist, prompting the VLM to generate key multiple questions to detect whether a target image is an anomaly. This method decomposes anomaly detection into multiple focused questions instead of relying on a single query. Recent studies \citep{ko-etal-2024-hierarchical, yang-etal-2024-decompose} show that task deconstruction methods improve reliability. Hence, our method makes judgements by integrating multiple main questions (Main-Qs). 

We provide the former summary and normality definition as input when prompting the VLM to extract key questions. The normality definition is reintroduced to help the VLM extract more relevant normality criteria. The resulting questions serve as candidate Main-Qs. Since only a few normal image descriptions are available, the initial set of questions may not fully generalize across all cases. To improve robustness, we evaluate their consistency by applying them to a diverse set of normal images. As questions with low accuracy (below 80\%) are indicative of bias toward the few-shot samples, they were excluded to ensure that the final set of questions remains broadly applicable without dataset-specific bias.

\subsection{Testing} 
In the testing step, the goal is to judge whether the query image is anomalous and to analyze the cause of the anomaly. Recent VLMs are not always reliable and may generate incorrect answers or suffer from hallucinations \citep{mashrur2024robust, zhang2024why}. To mitigate this, we augment each Main-Q with five semantically equivalent sub-questions (Sub-Qs) \citep{zhou2022large}. The final decision is made through majority voting on the Sub-Qs' responses.

By leveraging multiple outputs instead of a single response, our method effectively reduces reasoning errors. If any Main-Q receives a ‘No’ response, it means that the image violates at least one normal constraint and is classified as an anomaly. Additionally, the specific Main-Qs receiving ‘No’ provide a clear rationale for the anomaly’s cause.

To enhance interpretability, our approach follows a step-by-step \citep{kojima2022large} reasoning process rather than a direct anomaly prediction. This aligns with the CoT approach \citep{wei2022chain}, which strengthens VLM’s logical reasoning and maintains contextual consistency, thereby improving judgment reliability.

Unlike traditional AD methods that require class-specific prompts, LogicQA eliminates such dependencies, enabling flexible and intuitive modifications by adjusting only the question and class name \citep{portillo-wightman-etal-2023-strength}. This makes it highly applicable for industrial use, as it does not require predefined class-specific guided prompts or CoT reasoning like \citet{jin2025logicad}, allowing for seamless adoption in real-world settings.

\section{Dataset}
We evaluated our method using the MVTec LOCO AD dataset and an industrial semiconductor SEM dataset collected from real-world manufacturing processes. Both datasets contain normal and logical anomaly samples. ( The overview and sample images of the two datasets are included in the Appendix \ref{sec:loco_dataaset_overview} and \ref{sec:sem_dataset}.)  

\paragraph{MVTec LOCO AD Dataset} MVTec LOCO AD Dataset, (\citet{bergmann2022beyond}), consists of five object categories (breakfast box, juice bottle, pushpins, screw bag, splicing connectors) from industrial scenarios, with objects selected as close as possible to real-world applications. Each category has several types of logical anomaly. 

\begin{figure}[h!]
  \includegraphics[width=\columnwidth]{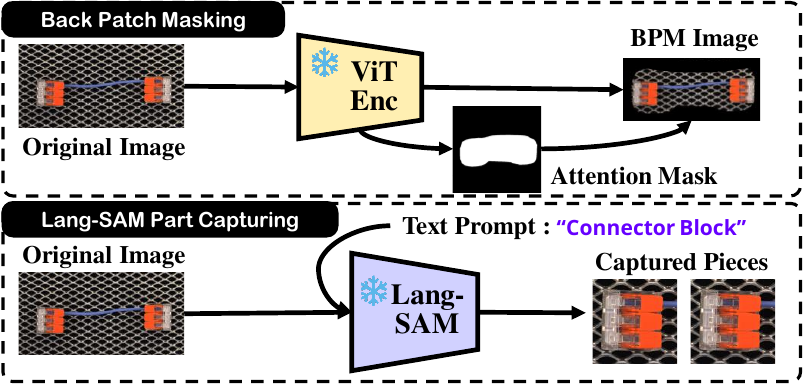}
  \caption{\textbf{Input Image Pre-Processing}: BPM applies an attention mask to the original image, masking the background, preserving objects. Lang-SAM identifies objects relevant to the given prompt and returns them as bounding boxes. }
  \label{fig:bpm_langsam}
\end{figure}

The VLM struggles with cases in the MVTec LOCO AD dataset where images contain large background areas, leading to long input contexts \citep{liu-etal-2024-lost}, or where they contain uniform objects \citep{campbell2024understanding}. To address this, we applied two pre-processing steps, as depicted in Figure~\ref{fig:bpm_langsam}. First, \textbf{Back Patch Masking (BPM)} \citep{lee2023uniformaly} was used to isolate the target object from the background, producing an object-centered image. Second, \textbf{Language Segment-Anything model (Lang-SAM)}, combined with GroundingDINO \citep{liu2024grounding} and SAM2 \citep{ravi2024sam}, was used to segment uniform objects individually, mitigating the VLM’s limitations in multi-object recognition. Details and effects of BPM and Lang-SAM are in the Appendix~\ref{sec:bpm_langsam} .

\paragraph{Semiconductor SEM Dataset} Scanning Electron Microscopy (SEM) operates by applying a high voltage to direct an electron beam onto the surface of a sample, then secondary electrons generate a wafer image. The SEM has around 1 nm resolution to get precise wafer surface patterns. This corporate dataset reflects critical inspection stages in semiconductor manufacturing, directly affecting chip quality and production yields. The dataset has two defect types: spot and bridge. Spot defects appear as circular blemishes that degrade chip performance, while bridge defects take the form of elongated connections linking separate conductive lines \citep{9200489}.

\begin{table*}[h!]
\resizebox{\textwidth}{!}{
\begin{tabular}{@{}cccccccccc@{}}
\toprule
\rowcolor{gray!15}
\begin{tabular}[c]{@{}c@{}} MVTec LOCO AD \\ (only Logical Anomaly)\end{tabular}    & \multicolumn{2}{c}{\textbf{LogicQA (Ours)}} & \multicolumn{2}{c}{\begin{tabular}[c]{@{}c@{}} LogicAD\\ \citet{jin2025logicad}\end{tabular}}   & \multicolumn{2}{c}{\begin{tabular}[c]{@{}c@{}} WinCLIP \\ \citet{jeong2023winclip}\end{tabular}  }  & \begin{tabular}[c]{@{}c@{}}PatchCore \\ \citet{roth2022towards} \end{tabular}   & \begin{tabular}[c]{@{}c@{}}GCAD\\ \citet{bergmann2022beyond}\end{tabular}  & \begin{tabular}[c]{@{}c@{}}AST\\ \citet{rudolph2023asymmetric} \end{tabular} 
\\ \midrule
Few / One shot  & \multicolumn{2}{c}{\textcolor{ForestGreen}{{\faCheck}}} & \multicolumn{2}{c}{\textcolor{ForestGreen}{{\faCheck}}} & \multicolumn{2}{c}{\textcolor{ForestGreen}{{\faCheck}}} & \textcolor{ForestGreen}{{\faCheck}}  & \faTimes & \faTimes \\ 
Explainable                       & \multicolumn{2}{c}{\textcolor{ForestGreen}{{\faCheck}}} & \multicolumn{2}{c}{\textcolor{ForestGreen}{{\faCheck}}} & \multicolumn{2}{c}{\faTimes} & \faTimes & \faTimes & \faTimes   \\ 
Auto-Generated Prompt & \multicolumn{2}{c}{\textcolor{ForestGreen}{{\faCheck}}} & \multicolumn{2}{c}{\faTimes} & \multicolumn{2}{c}{\faTimes} & \faTimes & \faTimes & \faTimes   \\ 
\midrule
\rowcolor{gray!15} Category & AUROC & $F_1$-max & AUROC & $F_1$-max & AUROC & $F_1$-max & AUROC & AUROC & AUROC \\  \midrule
Breakfast Box & 87.6 & \textbf{91.6} & \textbf{93.1} & 82.7 & 57.6 & 63.3 & 74.8  & 87.0 & 80.0\\ 
Juice Bottle & 88.2 & \textbf{89.6} & 81.6 & 83.2 & 75.1 & 58.2 & \textbf{93.9}  & 100.0  &   91.6  \\
Pushpins & \textbf{98.4} & 97.6 & 98.1 & \textbf{98.5} & 54.9 & 57.3 & 63.6  & 97.5  &   65.1  \\
Screw Bag & 71.5 & 64.5 & \textbf{83.8} & \textbf{77.9} & 69.5 & 58.8 & 57.8  & 56.0  &   80.1  \\
Splicing Connectors & \textbf{92.4} & \textbf{91.5} & 73.4 & 76.1 & 64.5 & 59.9 & 79.2  & 89.7  &   81.8  \\\midrule
\textbf{Average} & $\mathbf{87.6}\ (\mathbf{\textcolor{red}{1.6 \uparrow}})$ & $\mathbf{87.0}\ (\mathbf{\textcolor{red}{3.3 \uparrow}})$ & 86.0 & 83.7 & 64.3 & 59.5   & 74.0  & 86.0  &   79.7  \\ 
\bottomrule
  \end{tabular}
}
\caption{\textbf{Logical AD performance on MVTec LOCO AD dataset.} AUROC and $F_1$-max  in \% for detecting logical anomalies of all categories of MVTec LOCO AD Dataset. We report the mean over 3 runs for our method. Among models using the few-shot approach, the best results are highlighted in bold. The values highlighted in red indicate increased score compared to LogicAD. Our LogicQA demonstrates outstanding performance while incorporating a few-shot approach, explainability, and the use of auto-generated prompts.}
\label{tab:final_result_table}
\end{table*}

\section{Experiments and Results}
\subsection{Experimental Setting}
We implement our experiments by leveraging three SOTA VLMs (GPT-4o \citep{achiam2023gpt}, Gemini-1.5 Flash \citep{team2024gemini}, and InternVL-2.5 38B \citep{chen2024expanding}). Comprehensive details on model configurations and deployment settings are outlined in the Appendix~\ref{sec:implement_vlm}. All experiments are training-free and few-shot (three normal images per test image). Our assessments are based on the MVTec LOCO AD dataset and Semiconductor SEM dataset. We conducted the experiments three times for each category and calculated the average score, as indicated in Table~\ref{tab:final_result_table}.

\subsection{Evaluation Metrics}
Our approach uses a VLM for Vision Question-Answering \citep{sinha-etal-2025-guiding}. If any of the responses to Main-Qs are “No”, the model predicts “Anomaly”. It is threshold-free, providing binary predictions and reasoning but not an anomaly score. So, we propose using the VLM's log probabilities to compute an anomaly score. \citet{kadavath2022language, kim-etal-2024-probgate, lee-etal-2021-towards} have shown that low token prediction probabilities (\textit{Log probs}) can indicate a lack of knowledge in LLMs and lead to uncertain performance on downstream tasks. We consider the VLM's log-probability of answers to Sub-Qs as indicators of accuracy, reliability, and confidence of answer. We define key formulations:

A Sub-Q function \( q_{ij} \) outputs "Yes(0)" or "No(1)" for an input image \( x \), where \( i \in [1, m] \) represents the number of Main-Qs, and \( j \in [1,5] \) indexes the five Sub-Qs per Main-Q. Each Main-Q, \( Q_i(x) \) is defined as:
{\small
\[
Q_i(x) =
\begin{cases}
0  ,  & \text{if} \; \displaystyle\sum_{j=1}^{5} q_{ij}(x) < \displaystyle\sum_{j=1}^{5} (1 - q_{ij}(x)) \\
1  ,  & \text{otherwise}.
\end{cases}
\]
}
A final function \( F(x) \) determines whether the input is a normal image or an anomaly, defined as:

{\small
\[
F(x) =
\begin{cases}
\text{"Normal"} ,  & \text{if} \; \displaystyle\sum_{i} Q_i(x) = 0, \\
\text{"Anomaly"} , & \text{otherwise}.
\end{cases}
\]
}

For each Main-Q, we take the highest log-probability among the Sub-Qs whose answers match the voted result, then apply the exponential function to all selected values. And, we get anomaly score for test image below:
{\small
\begin{gather*}
s_i = \max_{j} \left\{ \log p(q_{ij}(x)) \mid q_{ij}(x) = Q_i(x) \right\} \\
S = \left\{ e^{s_i} \mid i = 1, \dots, m \right\} \\
\text{Anomaly Score} =
\begin{cases}
1 - \operatorname{Median}(S), & \text{if } F(x) \text{ = Normal}  \\
\operatorname{Median}(S), & \text{if } F(x) \text{ = Anomaly} 
\end{cases}
\end{gather*}
}

The \textit{logp} function computes the log probability generated during the processing of the input. By calculating the anomaly score as above, we use  $F_1$-max and Area Under the Reciever Operating Characteristic (AUROC) to evaluate our method, \textbf{LogicQA}, as same as existing approaches.

\subsection{Result}
\paragraph{MVTec LOCO AD Result}
The performance of Logical AD tested on the MVTec LOCO AD dataset for each method is shown in Table~\ref{tab:final_result_table}, presented in terms of AUROC and $F_1$-max scores. For a comprehensive comparison, the table also indicates which shot approach was chosen and whether explainability is incorporated. LogicQA consistently outperforms the existing few-shot VLM-based SOTA method \citep{jin2025logicad} across all metrics, achieving a 1.6\% increase in AUROC and a 3.3\% improvement in $F_1$-max score. Notably, in the \textit{splicing connectors} class, both the AUROC and $F_1$-max metrics showed remarkable improvements, with AUROC increasing by 19\% and $F_1$-max improving by 15.4\%. Even compared to full-shot methods \citep{liu2025gcad, rudolph2023asymmetric}, our LogicQA outperforms in almost all classes. (Frameworks utilizing in-house annotations are in Appendix~\ref{tab:full_final_result_table}). \\
LogicQA not only employs a few-shot approach and an auto-generated question mechanism for prediction but also provides natural language explanations for anomaly causes while achieving remarkable performance compared to other models.

\paragraph{Semiconductor SEM Result}
As shown in Table~\ref{tab:sem_result}, LogicQA (GPT-4o) outperforms PatchCore \citep{roth2022towards}, a representative few-shot AD method, on the semiconductor SEM dataset, yielding an 11.1\% increase in AUROC and a 14.6\% improvement in $F_1$-max. Also, LogicQA (GPT-4o) excels in detecting both “Bridge” and “Spot” anomalies, achieving the best scores. LogicQA significantly outperforms PatchCore even using the smaller open-source model InternVL-2.5 8B \citep{chen2024expanding}. This suggests applicability in real-world industrial settings, where deploying large proprietary models may not be feasible. Additionally, LogicQA shows excellent performance in Table~\ref{tab:sem_result} even though it did not include the process of filtering Main-Q using a few normal images.

\begin{table}[h!]
\resizebox{\columnwidth}{!}{
\begin{tabular}{@{}cccccc@{}}
\toprule
\multirow{3}{*}{SEM} &  \multicolumn{3}{c}{\textbf{LogicQA} }   & \multicolumn{2}{c}{PatchCore}  \\
  &  \multicolumn{2}{c}{ GPT-4o } & InternVL-2.5 8B  & \multicolumn{2}{c}{ \citet{roth2022towards} }    \\
\cmidrule(lr){2-6} & AUROC & $F_1$-max & $F_1$-max  & AUROC & $F_1$-max   \\ \midrule
 Bridge    & \textbf{89.7} & \textbf{90.4}  & 80.7 & 83.0 & 76.4  \\
Spot    & \textbf{90.8} & \textbf{94.3} & 89.7  & 75.4 & 79.2  \\ \midrule
Average & $\mathbf{90.3} \, (\mathbf{\textcolor{red}{11.1 \uparrow}})$ & $\mathbf{92.4} \, (\mathbf{\textcolor{red}{14.6 \uparrow}})$ & 85.2 & 79.2 & 77.8 \\ \bottomrule
\end{tabular}
}
\caption{\textbf{Logical AD performance on Semiconductor SEM dataset.} Our LogicQA outperforms PatchCore regarding metrics and AD explainability. All experiments were conducted with the same three normal images.}
\label{tab:sem_result}
\end{table}

\subsection{Ablation Studies}
 \paragraph{\textbf{\textit{Does LogicQA provide the correct reasoning?}}}
 The MVTec LOCO AD dataset does not provide specific rerasons for why each anomaly image is classified as anomalous. Therefore, we conducted a human evaluation to compare the reasons behind the model's anomaly detection with human perception.
 Two annotators were provided with the dataset and Main-Qs for each class and asked to answer accordingly. Their responses were then compared with the model’s answers. Annotator1 showed 98\% agreement for normal images and 85\% for anomalous ones, while Annotator2 showed 98\% and 86\%, respectively, demonstrating high correspondence. Notably, the strong agreement for anomalous images indicates that LogicQA not only detects anomalies but also explains their critical causes, demonstrating its ability as a comprehensive anomaly explainability model.

\paragraph{\textbf{\textit{Can other VLMs work well with LogicQA?}}} To verify the applicability of our LogicQA in other VLMs with fewer parameters, we conducted tests using Gemini-1.5 Flash \citep{team2024gemini} and InternVL-2.5 38B \citep{chen2024expanding}. The experimental results, presented in Table~\ref{tab:other_vlms} with recorded $F_1$-max scores, show that both models maintained stable performance, with some classes even achieving higher scores. This suggests that LogicQA can be effectively applied across various VLMs.

\begin{table}[h!]
\resizebox{\columnwidth}{!}{
\begin{tabular}{@{}cccc@{}}
\toprule
\textbf{VLMs} & GPT-4o & Gemini-1.5 Flash & InternVL-2.5 38B  \\ \midrule
 Breakfast Box &  91.6   & 83.3 &   88.2 \\
 Juice Bottle  & 89.6  &  78.0 &  73.7  \\
 Pushpins & 97.6 &  98.9  &  93.7   \\
Screw Bag & 64.5 & 91.7 & 62.6   \\
Splicing Connectors  & 91.5   & 46.8 & 69.9  \\ \midrule
Average  & 87.0   & 79.7 & 77.6 \\\bottomrule
\end{tabular}
}
\caption{\textbf{LogicQA performance with other VLMs on the MVTec LOCO AD dataset.} }
\label{tab:other_vlms}
\end{table}

\section{Conclusion}
In this paper, we propose LogicQA, an explainable logical AD framework leveraging a Vision-Language Model (VLM) to detect anomalies and provide natural language explanations. LogicQA requires only a few normal images to define normal characteristics, significantly reducing the dependency on large labeled datasets. By eliminating class-specific fine-tuning and manually generated prompts, LogicQA facilitates efficient and scalable deployment in industrial environments. We evaluated LogicQA on the public benchmark, MVTec LOCO AD Dataset, where it outperformed existing explainable AD models. We further validated robustness of LogicQA on a real-world manufacturing dataset, Semiconductor SEM Dataset. These results confirm LogicQA as an effective, reliable, and practical solution for diverse industrial applications.

\section*{Limitations}
Our framework is designed for easy application in industrial settings and delivers strong performance, though some limitations remain. Since our approach relies on VLMs, its performance inherently depends on the VLMs’ visual recognition capabilities. Currently, VLMs exhibit imperfect accuracy \citep{wang2023evaluation, li-etal-2023-evaluating} necessitating specific image preprocessing steps. However, as the technology evolves, this step may become less necessary \citep{jiang2025interpreting, liu2025reducing}.
Additionally, generating a well-generalized Main-Qs set requires diverse images. Fortunately, normal images are relatively easy to obtain in industrial environments \citep{9523565, liu2024deep}, which helps mitigate this challenge. Also, the evaluation result on the Semiconductor SEM dataset confirms our model demonstrated strong anomaly detection performance even without the Main-Q filtering process.

\section*{Ethics Statement}
This research uses GPT-4o and Gemini-1.5-Flash as baseline models. As with any large language model, their outputs may include unintended biases or harmful content depending on user inputs. To ensure ethical deployment, we apply engineering measures to mitigate these risks and enhance model reliability. Since both models are proprietary, with undisclosed training details and weights, assessing potential biases and risks remains challenging. Additionally, handling sensitive data with these models requires caution due to possible unintended exposure. When necessary, we recommend using open-source alternatives for greater transparency and control. AI-assisted tools were utilized solely for grammar correction and linguistic refinement during manuscript preparation. However, the originality, intellectual contributions, and core ideas of this paper are entirely the authors' own. We are committed to responsible AI use, continuous monitoring, and improving fairness and safety in real-world applications.


\bibliography{custom}

\appendix

\onecolumn

\section{LogicQA - Prompts} \label{sec:logicqa_prompt}

\begin{tcolorbox}[
    colback=white,
    colframe=black!70,
    arc=5pt,
    boxrule=1pt,
    title={\textcolor{black}{\textbf{Prompt - Describing the Normal Images}}},
    coltitle=black,
    fonttitle=\large,
    colbacktitle=JungleGreen!20,
]

\textbf{This is a \textcolor{Maroon}{\{Class\}}.}  
\textbf{Analyze the image and describe the \textcolor{Maroon}{\{Class\}} in detail, including type, color, size (length, width), material, composition, quantity, relative location.} \\ \\
\textbf{< Normal Constraints for a \textcolor{Maroon}{\{Class\}} >} \\
\textbf{\textcolor{Maroon}{\{Normal Definition\}}} \\ 

\textbf{\textcolor{Maroon}{\{Image Prompt (Image Input)\}}} \\
\hrule \;
\\
\textit{\textcolor{darkgray}{ Example } :}  \\ 
  \textit{\textcolor{darkgray}   {  This is a breakfast box. }} 
  \textit{  \textcolor{darkgray} {Analyze the image and describe the breakfast box in detail, including type, color, size (length, width), material, composition, quantity, relative location.}}. \\ \\
\textit{ \textcolor{darkgray}{    <Normal Constraints for breakfast box>}} \\
\textit{ \textcolor{darkgray} {  - The breakfast box always contain exactly  two tangerines and one nectarine that are always located on the left-hand side of the box.\\
  \; - The ratio and relative position of the cereals and the mix of banana chips and almonds on the right-hand side are fixed.}}

\end{tcolorbox}

\begin{tcolorbox}[
    colback=white,
    colframe=black!70,
    arc=5pt,
    boxrule=1pt,
    title={\textcolor{black}{\textbf{Prompt - Summarizing the Normal Image Context}}},
    coltitle=black,
    fonttitle=\large,
    colbacktitle=Cyan!20,
]
\textbf{[ Normal} \textbf{\textcolor{Maroon}{\{Class\}}  Description 1 ]} \\
\textbf{\textcolor{Maroon}{\{Description 1\}}} \\
\\
\textbf{[ Normal} \textbf{\textcolor{Maroon}{\{Class\}} Description 2 ]} \\
\textbf{\textcolor{Maroon}{\{Description 2\}}} \\
\\
\textbf{[ Normal }\textbf{\textcolor{Maroon}{\{Class\}} Description 3 ]} \\
\textbf{\textcolor{Maroon}{\{Description 3\}}} \\
\; \\
\textbf{Combine the three descriptions into one by extracting only the "common" features.} \\
\textbf{Create a concise summary that reflects the shared characteristics while removing any redundant or unique details.} \\
\hrule \;
\\
\textit{\textcolor{darkgray}{ Example } :}  \\  
\textit{\textcolor{darkgray}{ [ Normal Breakfast Box Description 1 ] }}  \\  
\textit{\textcolor{darkgray}{ The breakfast box is divided into two sections. ... } }  \\   \\ 
\textit{\textcolor{darkgray}{ [ Normal Breakfast Box Description 2 ] }}  \\  
\textit{\textcolor{darkgray}{ The breakfast box in the image contains the following items:. ... } }  \\ \\ 
\textit{\textcolor{darkgray}{ [ Normal Breakfast Box Description 3 ] }}  \\  
\textit{\textcolor{darkgray}{ The breakfast box in the image has two side. ... } }  \\ \\ 
\textit{\textcolor{darkgray}{ Combine the three descriptions into one by extracting only the "common" features. }  }  \\  
\textit{\textcolor{darkgray}{ Create a concise summary that reflects the shared characteristics while removing any
redundant or unique details. }  }    
\end{tcolorbox}

\begin{tcolorbox}[
    colback=white,
    colframe=black!70,
    arc=5pt,
    boxrule=1pt,
    title={\textcolor{black}{\textbf{Prompt - Generating Main Questions}}},
    coltitle=white,
    fonttitle=\large,
    colbacktitle=Plum!20,
]
\textbf{[ Description of } \textbf{\textcolor{Maroon}{\{Class\}} ]} \\
\textbf{\textcolor{Maroon}{\{ Summary Description \}}} \\
\\\textbf{[ Normal Constraints for } \textbf{\textcolor{Maroon}{\{Class\}} ]} \\
\textbf{\textcolor{Maroon}{\{Normal Definition\}}} \\ 

Using the \textbf{[ Normal Constraints for } \textbf{\textcolor{Maroon}{\{Class\}} ]} and \textbf{[ Description of } \textbf{\textcolor{Maroon}{\{Class\}} ]}, create several but essential , simple and important questions to determine whether the \textbf{\textcolor{Maroon}{\{Class\}} ]} in the image is normal or abnormal. 
Ensure the questions are only based on visible characteristics, excluding any aspects that cannot be determined from the image.
Also, simplify any difficult terms into easy-to-understand questions. \\
(Q1) : ...\\
(Q2) : ... \\
\hrule \;
\\\textit{\textcolor{darkgray}{ Example } :}  \\  
\textit{\textcolor{darkgray}{ [ Description of breakfast box ] }}  \\  
\textit{\textcolor{darkgray}{ The breakfast box is divided into two sections: ... }}  \\  

\textit{\textcolor{darkgray}{ [ Normal Constraints for breakfast box ] }}  \\ 
\textit{ \textcolor{darkgray} {  - The breakfast box always contain exactly  two tangerines and one nectarine that are always located on the left-hand side of the box.\\
  \; - The ratio and relative position of the cereals and the mix of banana chips and almonds on the right-hand side are fixed.}}
\\

\textit{\textcolor{darkgray}{ Using the [Normal Constriants for Breakfast Box] and [Description of Breakfast Box], create several but essential , simple and important questions to determine whether the Breakfast Box in the image is normal or abnormal. Ensure the questions are only based on visible characteristics, excluding any aspects that cannot be determined from the image. Also, simplify any difficult terms into
easy-to-understand questions.}}  \\ 
\textit{\textcolor{darkgray}{(Q1): ...}}  \\ 
\textit{\textcolor{darkgray}{(Q1): ...}}  
\end{tcolorbox}

\begin{tcolorbox}[
    colback=white,
    colframe=black!70,
    arc=5pt,
    boxrule=1pt,
    title={\textcolor{black}{\textbf{Prompt - Generating 5 variations Sub-Questions}}},
    coltitle=white,
    fonttitle=\large,
    colbacktitle=Lavender!20,
]
\textbf{ Generate five variations of the following question while keeping the semantic meaning. } \\
\textbf{ Input : } \textbf{\textcolor{Maroon}{\{Question\}} } \\
\textbf{ Output1: } \\
\textbf{ Output2: } \\
\textbf{ Output3: } \\
\textbf{ Output4: } \\
\textbf{ Output5: } \\
\hrule \;
\\\textit{\textcolor{darkgray}{ Generate five variations of the following question while keeping the semantic meaning. } } 
\\\textit{\textcolor{darkgray}{ Input : Is there one nectarine visible on the left-hand side of the breakfast box? } } 
\\\textit{\textcolor{darkgray}{  Output 1: } }
\\\textit{\textcolor{darkgray}{  Output 2: } }
\\\textit{\textcolor{darkgray}{  Output 3: } }
\\\textit{\textcolor{darkgray}{ Output 4: } }
\\\textit{\textcolor{darkgray}{  Output 5: } }
\end{tcolorbox}

\begin{tcolorbox}[
    colback=white,
    colframe=black!70,
    arc=5pt,
    boxrule=1pt,
    title={\textcolor{black}{\textbf{Prompt - Testing}}},
    coltitle=white,
    fonttitle=\large,
    colbacktitle=Lavender!20,
]
\textbf{ Question : } \textbf{\textcolor{Maroon}{\{Question\}} } \\
\textbf{        At first, describe \textcolor{Maroon}{\{Class\}} image and then answer the question. \\
        Your response  must end with `- Result: Yes` or `- Result: No`. \\
        Let's think step by step.} \\

\textbf{\textcolor{Maroon}{\{Test Image Prompt (Test Image Input)\}}}
\\
        
\hrule \; 
\\\textit{\textcolor{darkgray}{ Question : Can you see a single nectarine on the left side of the breakfast box? } }
\\\textit{\textcolor{darkgray}{ At first, describe breakfast box image and then answer the question. \\ Your response must end with `- Result: Yes` or `- Result: No`. \\
        Let's think step by step. } }
\end{tcolorbox}
\; \\
\; \\

\section{VLM Implementation Details} \label{sec:implement_vlm}

\subsection{VLMs} 

In our study, we uses three VLMs: GPT-4o \citep{achiam2023gpt} , Gemini-1.5 Flash \citep{team2024gemini} , and InternVL\-2.5(38B, 8B) \citep{chen2024expanding}. The GPT-4o model was accessed and inferred through the OpenAI API. For the GPT-4o model, we fixed \textit{temperature} to 1.0 and other hyper-parameters to default. Regarding the Gemini-1.5  models, \,\textit{temperature} is 1, \,\textit{top\_p} is 0.95, \,and \textit{top\_k} is 40. For Open-Source InternVL-2.5 from OpenGVLab, we set \textit{temperature} to 0.2,  \,\textit{top\_p} to 0.7,\, \textit{repetition\_penalty} to 1.1, \, \textit{do\_sample} to True, and \, \textit{max\_new\_tokens} to 512.  \textbf{All these settings are the same across all experiments and across datasets.}

\subsection{Local Experimental Setup} 
We utilized the open-source InternVL-2.5, leveraging up to three NVIDIA A100 GPUs due to its substantial computational requirements.

\subsection{Lang-SAM Prompt}
When using Lang-SAM to the two classes \texttt{(Pushpins, Splicing Connectors)}, a text prompt was needed to accurately capture the independent entities. It is as follows.\\
- \texttt{Splicing Connectors: Connector Block} \\
- \texttt{Pushpins: The individual black compartments within the transparent plastic storage box}

\subsection{Data Security Option}
To ensure the confidentiality and security of the \textbf{Semiconductor SEM dataset} provided by global company, we took stringent precautions when utilizing GPT-4o for our research. \textcolor{red}{\textbf{Specifically, all data-sharing functionalities were disabled to strictly prevent unintended exposure or transmission of data outside the controlled research environment.}} By implementing these safeguards, we ensured that no proprietary or sensitive information was inadvertently shared with external servers or third-party entities. This approach aligns with best practices for handling proprietary industrial datasets while leveraging advanced AI models for research and analysis.

\newpage

\section{MVTec LOCO AD Dataset}
\label{sec:loco_dataaset_overview}
\subsection{MVTec LOCO AD Dataset Overview}

This is a statistical outline of the public MVTec Logical Constraints Anomaly Detection (LOCO) AD Dataset. It consists of five categories (\texttt{Breakfast Box, Screw Bag, Pushpins, Splicing Connectors, Juice Bottle}). We conducted a few-shot experiment by randomly selecting three photos from the train-normal set.

\begin{table}[h!]
\resizebox{\columnwidth}{!}{
\begin{tabular}{@{}ccccc@{}}
\toprule
Category & Train-Normal Images & Test-Normal Images & Test-Logical Anomaly Images & Detect types  \\ \midrule
 Breakfast Box & 351   & 102 &   83 & 22 \\
 Screw Bag  & 360  &  122 &  137 & 20  \\
 Pushpins & 372 &  138  &  91 & 8   \\
 Splicing Connectors  & 354   & 119 & 108 & 21 \\ 
 Juice Bottle & 335 & 94& 142 & 18   \\
\midrule
Total  & 1772   & 575 & 561 & 89 \\\bottomrule
\end{tabular}
}
\caption{\textbf{Overview of the MVTec LOCO AD dataset} }
\label{tab:loco_info}
\end{table}

\begin{figure}[htbp]
  \includegraphics[width=\columnwidth]{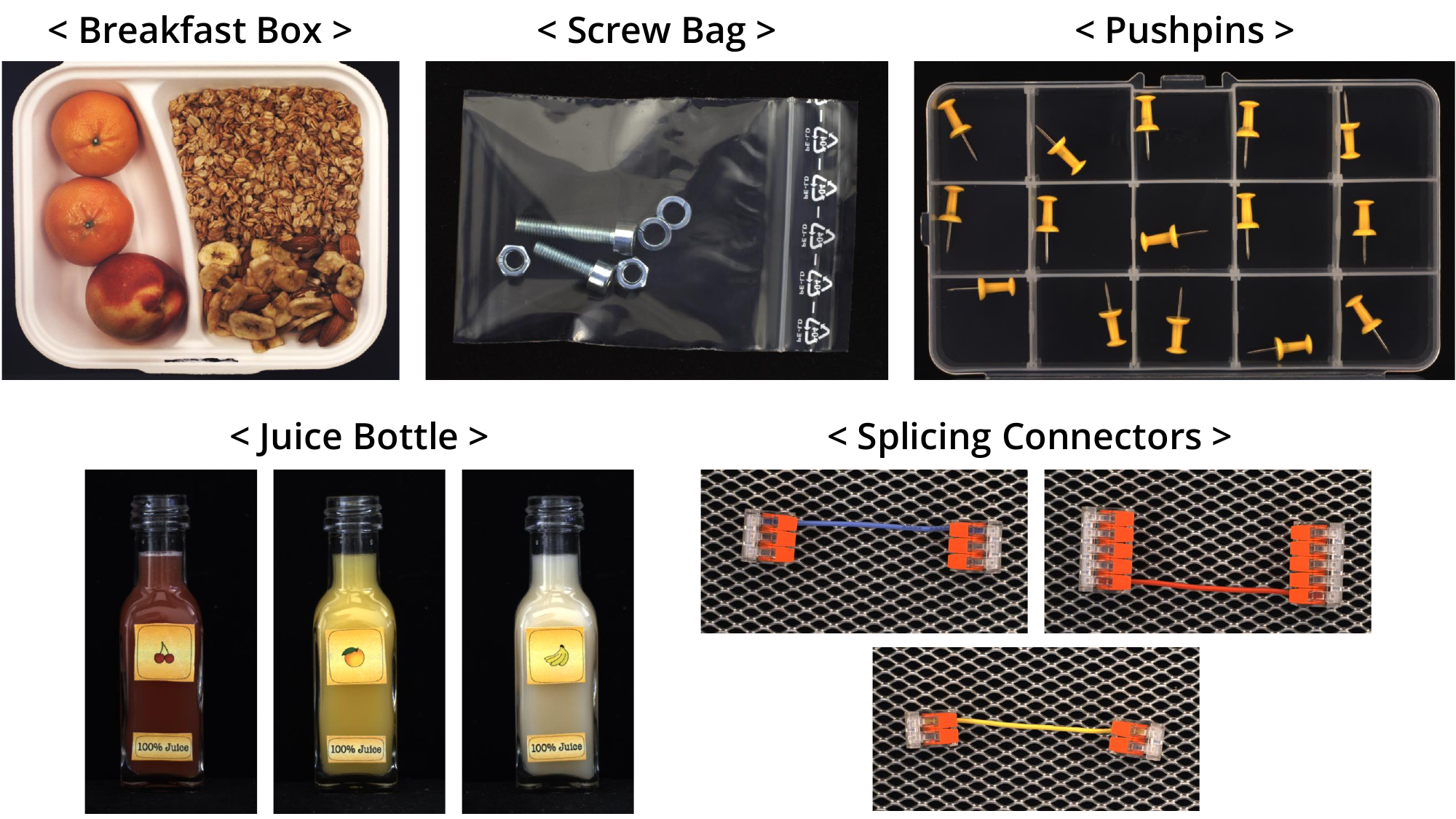}
  \caption{\textbf{MVTec LOCO AD Dataset Normal sample images}}
  \label{fig:Loco_ad_image}
\end{figure}

\subsection{ MVTec LOCO AD Dataset- Normality Definition for each class}
\label{subsec:normal_df}
 
Below is a summary of the normality definitions for each class. For \textit{Splicing Connectors} and \textit{Juice Bottle}, the normality definitions partially change depending on the color of each cable and the fruit of the juice. The changed parts are expressed in red.\\
\; \\
 \mybox{Breakfast Box}{yellow!30}{white!10}{
- The breakfast box always contain exactly  two tangerines and one nectarine that are always located on the left-hand side of the box. \\
- The ratio and relative position of the cereals and the mix of banana chips and almonds on the right-hand side are fixed.  }
 \mybox{Screw Bag}{yellow!30}{white!10}{
 - A screw bag contains exactly two washers, two nuts, one long screw, and one short screw.\\
- All bolts (screws) are longer than 3 times the diameter of the washer.} 
 \mybox{Pushpins}{yellow!30}{white!10}{
 - Each compartment of the box of pushpins contains exactly one pushpin.} 

 \mybox{Splicing Connectors}{yellow!30}{white!10}{- Exactly two splicing connectors with the same number of cable clamps are linked by exactly one cable. \\
- In addition, the number of clamps has a one-to-one correspondence to the \textcolor{Maroon}{\{color\}} of the cable. \\
- The cable must be connected to the same position on both connectors to maintain mirror symmetry. \\
- The cable length is roughly longer than the length of the splicing connector terminal block.}

\mybox{Juice Bottle}{yellow!30}{white!10}{
- The juice bottle is filled with   \textcolor{Maroon}{\{fruit\}}  juice and carries exactly two labels. \\
- The first label is attached to the center of the bottle, with the \textcolor{Maroon}{\{fruit\}} icon positioned exactly at the center of the label, clearly indicating the type of  \textcolor{Maroon}{\{fruit\}} juice.\\
- The second is attached to the lower part of the bottle with the text “100\% Juice” written on it. \\
- The fill level is the same for each bottle.\\
- The bottle is filled with at least 90\% of its capacity with juice, but not 100\%.  }

\subsection{Main-Questions for each class}

\mybox{Breakfast Box}{red!30}{white!10}{
Q1 : Are there exactly two tangerines visible on the left-hand side of the breakfast box? \\
Q2 : Is there one nectarine visible on the left-hand side of the breakfast box? \\
Q3 : Does the right-hand side of the breakfast box have cereals in the upper portion? \\
Q4 : Is there a mix of banana chips and almonds in the lower portion of the right-hand side of the breakfast box? \\
Q5 : Are the fruits (tangerines and nectarine) only on the left-hand side, and are the cereals with banana chips and almonds only on the right-hand side?}

\mybox{Screw Bag}{red!30}{white!10}{
Q1 : Are there exactly two tangerines visible on the left-hand side of the breakfast box? \\
Q2 : Is there one nectarine visible on the left-hand side of the breakfast box? \\
Q3 : Does the right-hand side of the breakfast box have cereals in the upper portion? \\
Q4 : Is there a mix of banana chips and almonds in the lower portion of the right-hand side of the breakfast box? \\
Q5 : Are the fruits (tangerines and nectarine) only on the left-hand side, and are the cereals with banana chips and almonds only on the right-hand side?}

\mybox{Pushpins}{red!30}{white!10}{

Q1 : Is there exactly one pushpin visible in the compartment? \\
Q2 : Is the pushpin yellow in color? \\
Q3 : Is the compartment transparent, allowing the pushpin to be visible? \\
Q4 : Is the pushpin visible against a contrasting background? }

\mybox{Splicing Connectors - Blue }{red!30}{white!10}{
Q1 : Are there exactly two splicing connectors visible in the image? \\
Q2 : Do both connectors have the same number of wire clamps? \\
Q3 : Is there only one blue cable connecting the two splicing connectors? \\
Q4 : Do the connectors have transparent bodies with orange levers? \\
Q5 : Do both connectors have three orange levers, indicating three cable clamps? \\
Q6 : Are the connectors made from clear plastic with metal contacts inside? \\
Q7 : Are the orange levers made of plastic? \\
Q8 : Is the blue cable connected to the same position on both connectors? \\
Q9 : Is the pushpin visible against a contrasting background? \\
Q10 : Does the blue cable appear longer than the length of one of the splicing connectors?}

\mybox{Splicing Connectors - Red }{red!30}{white!10}{
Q1 :  Are there exactly two splicing connectors in the image? \\
Q2 : Do both connectors have transparent casings with red or orange clamps/levers? \\
Q3 : Are the connectors rectangular and compact, each containing five clamps? \\
Q4 : Is there a single red cable connecting the two splicing connectors? \\
Q5 : Is the red cable slightly longer than the length of the splicing connector terminal block? \\
Q6 : Are the connectors positioned parallel to each other? \\
Q7 : Are the splicing connectors transparent with orange levers? \\
Q8 : Does the cable connect to the same clamp position on both connectors, maintaining mirror symmetry? \\
Q9 : Are the connectors made of plastic with transparent casings? }

\mybox{Splicing Connectors - Yellow }{red!30}{white!10}{
Q1 : Are there exactly two splicing connectors visible in the image?\\
Q2 : Do both splicing connectors have the same number of levers? \\
Q3 : Is the cable connecting the two splicing connectors yellow in color? \\
Q4 : Does each connector have two levers, indicating two clamps? \\
Q5 : Is the cable entering the same position on both connectors, maintaining symmetry? \\
Q6 :  Is the length of the yellow cable longer than the terminal block of each splicing connector? \\
Q7 : Are the splicing connectors transparent with orange levers?\\
Q8 : Are the connectors positioned symmetrically on either side of the yellow cable?\\
Q9 : Is there exactly one yellow cable connecting the two splicing connectors? }

\mybox{Juice Bottle - Orange }{red!30}{white!10}{
Q1 : Is the juice bottle filled with orange juice up to at least 90\% of its capacity, but not completely full? \\
Q2 : Are there exactly two labels on the juice bottle? \\
Q3 : Is the center label positioned in the middle of the bottle with an orange icon clearly visible? \\
Q4 : Does the center label have a light orange background? \\
Q5 : Is the lower label attached to the lower part of the bottle? \\
Q6 :  Does the lower label display the text 100\% Juice in bold, likely black, font? \\
Q7 :  Are the labels vertically aligned, with the center label above the lower label, creating a balanced appearance?  }

\mybox{Juice Bottle - Cherry }{red!30}{white!10}{
Q1 : Is the bottle made of clear glass, allowing the color of the cherry juice to be visible? \\
Q2 : Does the bottle have a central label with a cherry icon precisely placed in the middle?\\
Q3 : Is there a central label on the bottle with a cherry icon clearly indicating the type of juice? \\
Q4 : Is there a lower label on the bottle with the text 100\% Juice written on it? \\
Q5 : Is the fill level of the juice in the bottle at least 90\% of its capacity, with a small gap at the top indicating it is not completely full?  \\
Q6 : Is there a central label on the bottle with a cherry icon positioned exactly at the center of the label? \\
Q7 : Is the color of the juice a deep reddish-brown, consistent with cherry juice? }

\mybox{Juice Bottle - Banana }{red!30}{white!10}{
Q1 : Is the bottle made of clear glass, allowing you to see the banana juice inside?\\
Q2 :  Does the juice inside the bottle appear as a creamy, light yellow color, typical of banana juice? \\
Q3 : Is the bottle slender and of a standard size typically used for single-serve juice bottles? \\
Q4 :  Is there a central label on the bottle with a banana icon located exactly at the center of the label? \\
Q5 : Is there a lower label on the bottle that reads ??00\% Juice? \\
Q6 : Does the juice fill level reach at least 90\% of the bottle's capacity, with a small gap at the top? \\
Q7 : Are there exactly two labels on the bottle, one in the center and one lower down? }

\newpage

\subsection{Sub-Questions for each class}
An example of a sub-question configuration for the breakfast box class is given. The Sub-Questions can be created by applying an augmentation prompt (generating 5 variations Sub-Questions) to the Main-Questions. \\

\mybox{Breakfast Box}{red!30}{white!10}{
 \textbf{Q1 Sub-Questions}  
 
- Can you see exactly two tangerines on the left side of the breakfast box?  \\ 
-  Is the left-hand side of the breakfast box showing precisely two tangerines?  \\ 
- Do you observe exactly two tangerines on the left of the breakfast box? \\
- Are precisely two tangerines visible on the left side of the breakfast box? \\ 
- Does the left-hand side of the breakfast box contain exactly two tangerines?  \\ \; \\

 \textbf{Q2 Sub-Questions} 
 
- Can you see a single nectarine on the left side of the breakfast box? \\ 
- Is there a nectarine present on the left-hand side of the breakfast box? \\
- Do you spot one nectarine on the left area of the breakfast box? \\
- Is a nectarine visible on the left side within the breakfast box? \\
- Is there one nectarine that can be seen on the left part of the breakfast box? \\ \; \\

 \textbf{Q3 Sub-Questions}  
 
- Are there cereals located in the upper part of the right side of the breakfast box? \\
 - Is the upper portion of the right side of the breakfast box filled with cereals? \\
 - Can cereals be found in the top section on the right-hand side of the breakfast box? \\
 - Does the upper section of the right side of the breakfast box contain cereals? \\
 - Is the top of the right-hand side of the breakfast box occupied by cereals? \\ \; \\

 \textbf{Q4 Sub-Questions}  
 
 - Does the lower section on the right side of the breakfast box contain a combination of banana chips and almonds? \\
 - Can you find a blend of banana chips and almonds in the bottom part of the right-hand side of the breakfast box? \\
 - Are banana chips and almonds mixed together in the lower right section of the breakfast box? \\
 - Is there a combination of banana chips and almonds located in the bottom right area of the breakfast box? \\
 - Are banana chips and almonds present together in the lower portion on the right side of the breakfast box?'\\ \; \\

 \textbf{Q5 Sub-Questions}  
 
 -  Are tangerines and nectarines exclusively on the left, and are cereals with banana chips and almonds exclusively on the right? \\
 - Is it true that the fruits, such as tangerines and nectarines, are solely placed on the left while cereals with almonds and banana chips are only on the right? \\
 - Are the tangerines and nectarines located only on the left side, and are the cereals containing banana chips and almonds solely on the right side? \\
 - Are fruits like tangerines and nectarines restricted to the left-hand side, while cereals with banana chips and almonds are found only on the right? \\
 - Is the placement such that tangerines and nectarines are just on the left, and cereals with almonds and banana chips appear only on the right? 
}

\subsection{ Logical AD performance on MVTec LOCO AD dataset.}
\begin{table*}[h!]
\resizebox{\textwidth}{!}{
\begin{tabular}{@{}cccccccccccc}
\toprule
\begin{tabular}[c]{@{}c@{}}MVTec LOCO AD \\ (only Logical Anomaly)\end{tabular}    
& \multicolumn{2}{c}{\textbf{LogicQA} (Ours)} 
& \multicolumn{2}{c}{\begin{tabular}[c]{@{}c@{}} LogicAD\\ \citet{jin2025logicad}\end{tabular}}   
& \multicolumn{2}{c}{\begin{tabular}[c]{@{}c@{}} WinCLIP \\ \citet{jeong2023winclip}\end{tabular}}  
& \begin{tabular}[c]{@{}c@{}}PatchCore \\ \citet{roth2022towards} \end{tabular}   
& \begin{tabular}[c]{@{}c@{}}GCAD\\ \citet{bergmann2022beyond}\end{tabular} 
& \begin{tabular}[c]{@{}c@{}}AST\\ \citet{rudolph2023asymmetric} \end{tabular} 
& \begin{tabular}[c]{@{}c@{}}LogiCode\\ \citet{zhang2024logicode}\end{tabular} 
& \begin{tabular}[c]{@{}c@{}}PSAD\\ \citet{kim2024few} \end{tabular} 
\\ \midrule
Category & AUROC & $F_1$-max & AUROC & $F_1$-max & AUROC & $F_1$-max & AUROC & AUROC & AUROC& AUROC & AUROC \\  \midrule
Breakfast Box & 87.6 & 91.6 & 93.1 & 82.7 & 57.6 & 63.3 & 74.8  & 87.0 & 80.0 &98.8 & 100.0\\ 
Juice Bottle & 88.2 & 89.6 & 81.6 & 83.2 & 75.1 & 58.2 & 93.9  & 100.0  &   91.6 &99.4 &99.1 \\
Pushpins & 98.4 & 97.6 & 98.1 & 98.5 & 54.9 & 57.3 & 63.6  & 97.5  &   65.1 &98.8 &100.0 \\
Screw Bag & 71.5 & 64.5 & 83.8 & 77.9 & 69.5 & 58.8 & 57.8  & 56.0  &   80.1  &98.2 &99.3\\
Splicing Connectors & 92.4 & 91.5 & 73.4 & 76.1 & 64.5 & 59.9 & 79.2  & 89.7  &   81.8 &98.9 &91.9 \\\midrule
\textbf{Average} & $\mathbf{87.6}\ $                           & $\mathbf{87.0}\ $               & 86.0 & 83.7 & 64.3 & 59.5   & 74.0  & 86.0  &   79.7  &98.8 & 98.1\\ 
\bottomrule   \end{tabular}}
\caption{\textbf{(Extension Ver.) Logical AD performance on MVTec LOCO AD dataset.} AUROC and $F_1$-max  in \% for detecting logical anomalies of all categories of MVTec LOCO AD Dataset.}
\label{tab:full_final_result_table}
\end{table*}

\section{Can an Anomaly Score be effectively derived from the Token Prediction Probability?}

We propose using  VLM's Log Probabilities to compute an anomaly score. We assume that low token prediction probabilities (\textit{log\_probs}) lead to uncertain performance and incorrect answers, as in typical LLM studies. Therefore, we conducted additional experiments to verify whether this assumption is correct in our VLM task.

We extracted 50 normal images for each class and generated answers for each Main-Question. The VLM's answer must be "Yes" for all normal images. Therefore, if it is "No", the answer generated by VLM is incorrect. We visualized each answer and the average token prediction probability at that time by class.

\begin{figure}[htbp]
  \includegraphics[width=\columnwidth]{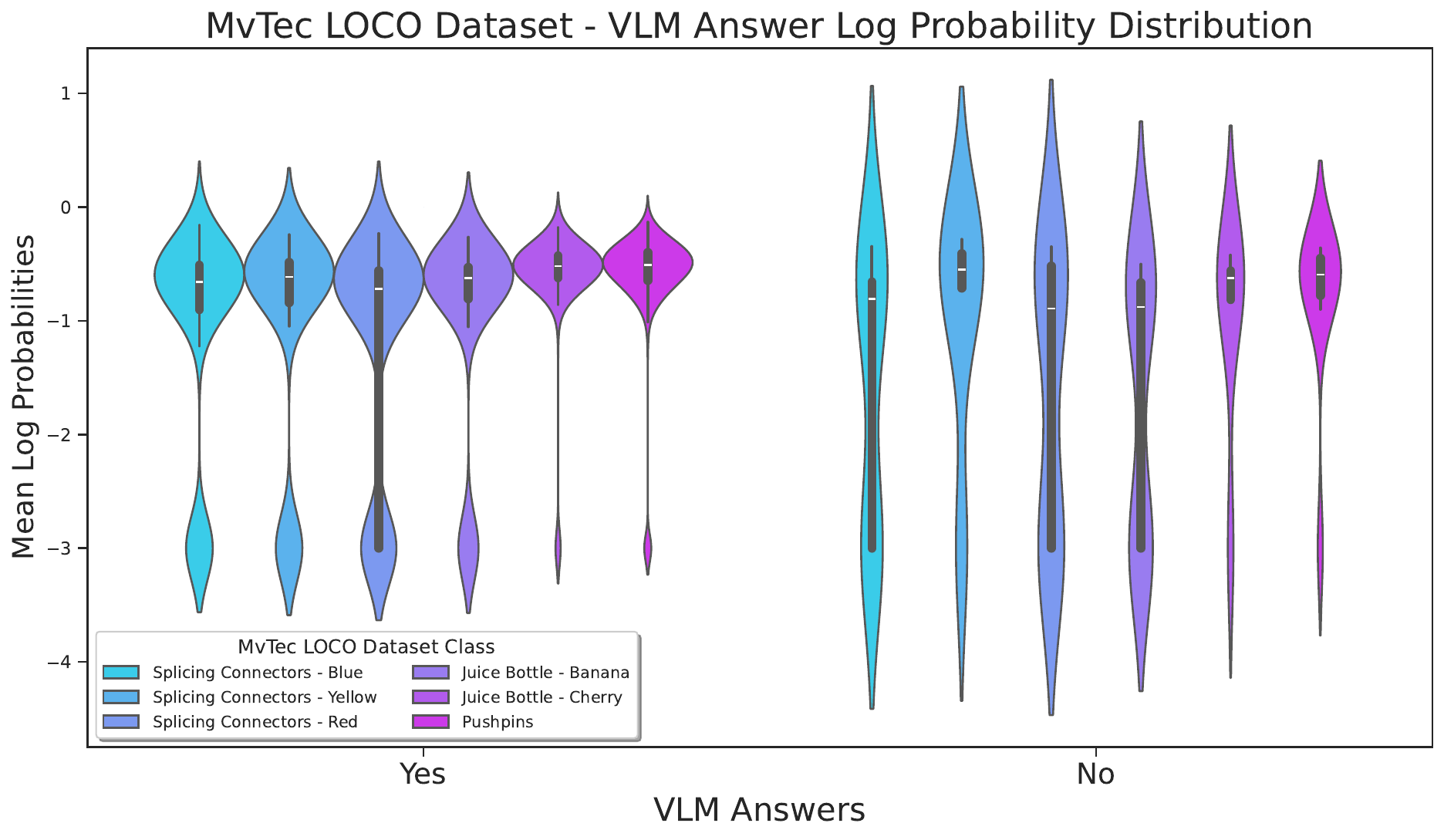}
  \caption{\textbf{Log-Probability Distribution of VLM answers}}
  \label{fig:violin_probs}
\end{figure}

As you can see from the figure~\ref{fig:violin_probs}, when generating the wrong answer "No" in some classes, the distribution of \textit{log\_probs} is generated relatively widely. When VLM generating "Yes", there is a clear section where the \textit{log\_probs} remains high, whereas in the case of "No", the \textit{log\_probs} come out quite diversely. Since our assumption is quite consistent with the actual data, it suggests that \textbf{as a result of verifying with actual data, it was confirmed that using the token prediction probability as the reliability of the answer and using it as the Anomaly Score is valid.}

\newpage

\section{Semicomductor SEM Dataset}
\label{sec:sem_dataset}

This is an overview of the Semiconductor SEM Dataset.
Scanning Electron Microscopy (SEM) operates by applying a high voltage to direct an electron beam onto the surface of a sample, then detecting secondary electrons that react to this beam to generate an image. The equipment used in our experiments achieves a resolution of approximately 1 nm, making it highly effective for observing the minute patterns on wafer surfaces.

Semiconductor fabrication involves hundreds to thousands of processing steps, comprising dozens of layers. Furthermore, each layer has a distinct pattern to form integrated circuits. This indicates a wide variety of both normal and abnormal (defective) patterns, implying that a generalized anomaly detection model would require an enormously large memory bank. 

There is two defect types for anomaly dataset, \texttt{Spot Defect} and \texttt{Bridge Defect}. These two types of anomaly sets share the same Normal dataset. Bridge defects occur when separate conductive lines or elements accidentally fuse, potentially causing short circuits. In contrast, spot defects appear as small, localized flaws on the wafer surface that can degrade overall device performance.

\textcolor{red}{The data was provided by a global semiconductor company, and the actual data cannot be disclosed for security reasons. The sample examples below are images similar to the actual images found in the paper~\citep{9200489} and attached.}

\begin{table}[h!]
\resizebox{\columnwidth}{!}{
\begin{tabular}{@{}ccccc@{}}
\toprule
Type & Train-Normal Images & Test-Normal Images & Test-Logical Anomaly Images   \\ 
\midrule
 Spot Defect  & \multirow{2}{*}{342}  & \multirow{2}{*}{169} &  290 \\
 Bridge Defect &   &     &  123   \\
\midrule
Total  & 342   & 169 &  413  \\\bottomrule
\end{tabular}
}
\caption{\textbf{Overview of the Semiconductor SEM dataset} }
\label{tab:sem_info}
\end{table}

\begin{figure}[htbp]
  \includegraphics[width=\columnwidth]{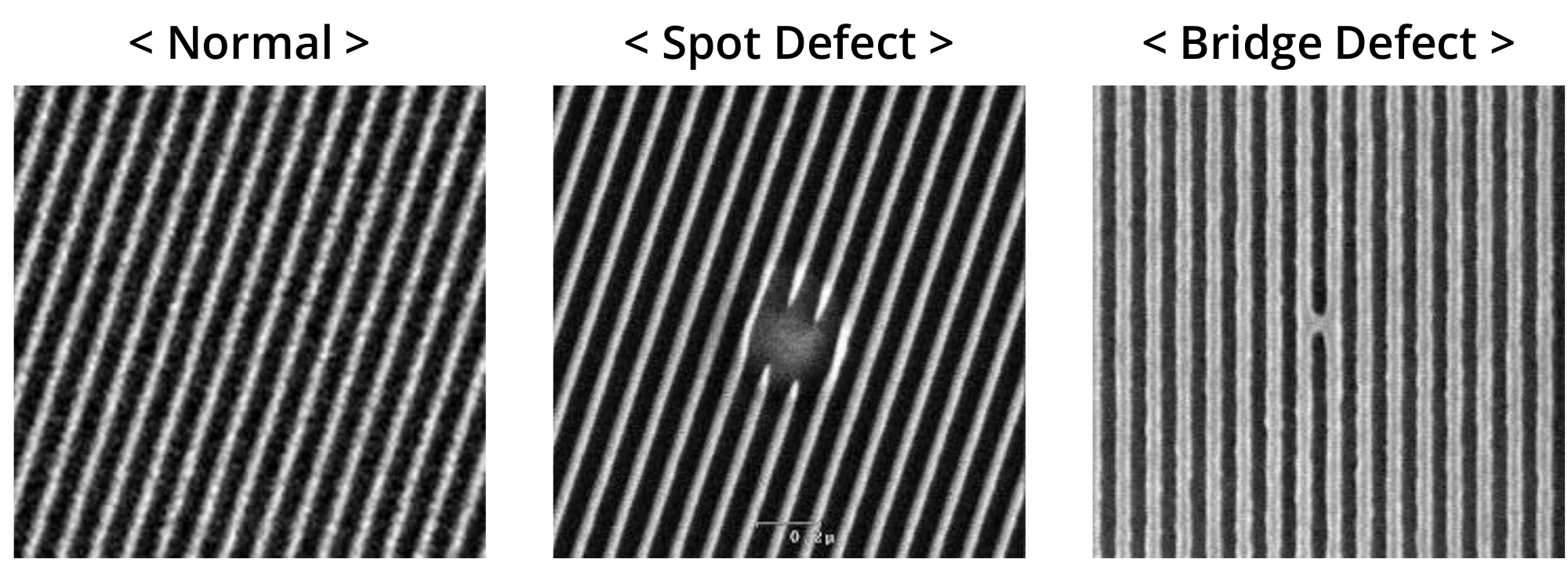}
  \caption{\textbf{Semiconductor SEM Dataset sample images}}
  \label{fig:sem_image}
\end{figure}

\subsection{ Semiconductor SEM Dataset- Normality Definition}

 \mybox{SEM wafer}{yellow!30}{white!10}{
- There should be no Particles, Hot Spots, or Defects.  }

\subsection{Main-Questions}

\mybox{SEM wafer}{red!30}{white!10}{
Q1 : Are there no visible particles or dust on the wafer surface? \\
Q2 : Are the etched patterns consistent and evenly spaced across the image? \\
Q3 : Is the surface free of bright or dark spots that look out of place? \\
Q4 : Do the etched lines appear smooth and uniform without breaks or distortions? \\
Q5 : Does the wafer surface look clean without any unexpected irregularities?}

\subsection{Sub-Questions}

\mybox{SEM wafer}{red!30}{white!10}{
 \textbf{Q1 Sub-Questions}  
 
 - Is the wafer surface completely free of visible particles or dust?  \\
 - Are there any visible particles or dust present on the wafer surface? \\
 - Can you confirm that no visible particles or dust are on the wafer surface? \\ 
 - Is the wafer surface entirely clean without any visible dust or particles? \\ 
 - Do you see any visible dust or particles on the wafer surface? \\ \; \\

 \textbf{Q2 Sub-Questions}  
 
 - Are the etched patterns uniform and evenly distributed throughout the image?  \\
 - Do the etched patterns appear consistent and evenly spaced across the entire image?  \\
 - Are the etched designs evenly spaced and consistent throughout the image?  \\ 
 - Is there uniformity in the etched patterns, with even spacing across the image?  \\ 
 - Do the etched patterns maintain consistency and equal spacing across the image? \\ \; \\

 \textbf{Q3 Sub-Questions}  
 
 - Does the surface have any unusual bright or dark spots? \\
 - Are there any bright or dark spots on the surface that seem out of place?   \\
 - Is the surface completely uniform, without any irregular bright or dark spots? \\ 
 - Do you notice any unexpected bright or dark spots on the surface? \\ 
 - Is the surface free from any abnormal bright or dark spots?   \\ \; \\

 \textbf{Q4 Sub-Questions}  
 
 - Are the etched lines consistently smooth and uniform, without any interruptions or distortions?  \\
 - Do the etched lines maintain a smooth and even appearance, free from breaks or irregularities?  \\
 - Are the etched lines free from distortions and interruptions, appearing smooth and uniform?  \\ 
 - Do the etched lines exhibit a continuous, smooth, and uniform pattern without any breaks?  \\ 
 - Are the etched lines well-defined, smooth, and uniform, without any visible distortions or gaps? \\ \; \\

 \textbf{Q5 Sub-Questions}  
 
 - Is the wafer surface free of any unexpected irregularities and appears clean?   \\
 - Does the wafer surface appear smooth and without any unwanted defects?  \\
 - Is the wafer surface visibly clean and devoid of any unexpected anomalies?  \\ 
 - Can you confirm that the wafer surface is clean and free from irregularities?  \\ 
 - Does the wafer surface exhibit a clean appearance without any noticeable defects? \\ \; \\
}

\newpage

\section{Details and Effect of BPM \& Lang-SAM} \label{sec:bpm_langsam}

The MVTec LOCO AD Dataset required image preprocessing based on class-specific features. In the \texttt{Splicing Connectors} class, the background consists of wire entanglement, while in the \texttt{Screw Bag} class, a large portion of the image is occupied by empty space within the bag. To address this, we applied \textbf{Back Patch Masking (BPM)} to these two classes. BPM isolates the foreground target from the background, enabling target-centric detection. Also, \texttt{Pushpins} class is uniformly placed in each compartment, and \texttt{Splicing Connectors} class consists of multiple identical terminals within each connector block. Since both classes exhibit the uniform objects issue that makes hallucination problem in VLM, we processed images using \textbf{Lang-SAM}. 

We conducted an experiment to verify whether BPM is actually effective in improving the response accuracy of VLM. We composed a subset of 50 normal images, entered the Main-Question for each class, and checked the answer. A normal image must answer "Yes" to the Main-Questions. If it answered a "No", VLM generated an wrong answer. We calculated the correct answer rate (accuracy) for each Main-Question for a total of 50 normal images. As you can see in the figure~\ref{fig:Lang_bpm_effect} below, \textbf{the accuracy of the answer increases when BPM is processed compared to when it is not.}

We also experimented to verify whether Lang-SAM is effective for VLM performance. We conducted an experiment with the same settings as the previous BPM additional experiment. As shown in figure~\ref{fig:Lang_bpm_effect}, we found that \textbf{Lang-SAM was significantly effective in improving the accuracy of VLM answers in both classes (Pushpins and Splicing Connectors).}
\\

\begin{figure}[htbp]
  \includegraphics[width=\columnwidth]{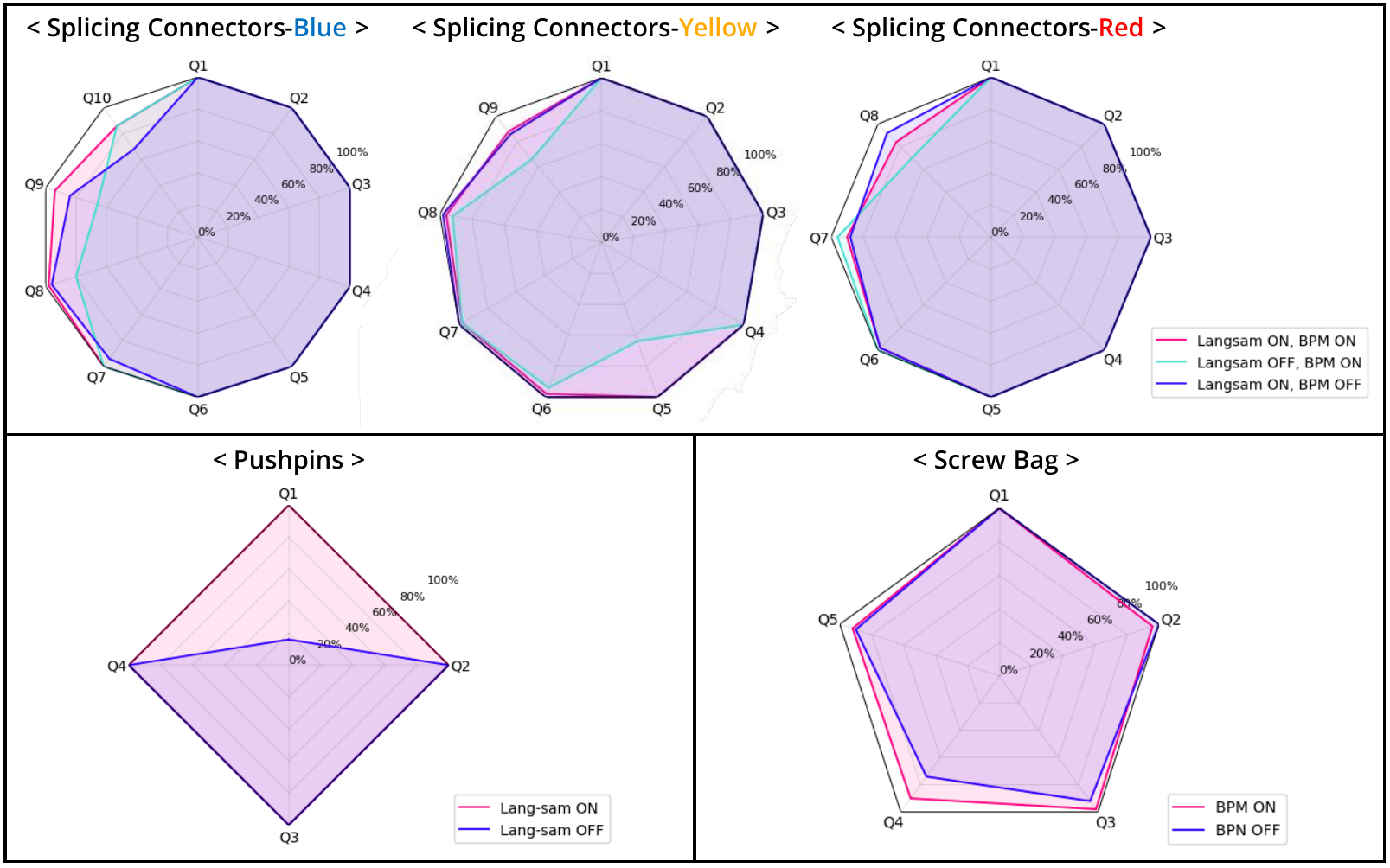}
  \caption{\textbf{BPM and Lang-SAM Effect for each class}}
  \label{fig:Lang_bpm_effect}
\end{figure}

\end{document}